\documentclass[journal,oneside,print]{ieeecolor}  
\usepackage{generic}
\usepackage{cite}
\usepackage{amsmath,amssymb,amsfonts}
\usepackage{algorithmic}
\usepackage{graphicx}
\usepackage{multirow}
\usepackage{array}
\usepackage{algorithm,algorithmic}
\usepackage{hyperref}
\hypersetup{hidelinks=true}
\usepackage{textcomp}
\usepackage[normalem]{ulem}

\usepackage{orcidlink}
\usepackage{subcaption}

\usepackage{etoolbox}
\makeatletter
\patchcmd{\thebibliography}{\section*{\color{black}}}{\section*{References}}{}{}
\makeatother

\newcommand{\w}{\textit{w/} }
\newcommand{\wo}{\textit{w/o} }
\newcommand{\SD}[1]{\small$\pm$ {#1}}

\definecolor{QGcolor}{RGB}{0,220,0}
\definecolor{QRcolor}{RGB}{255,0,0}
\definecolor{Ocolor}{RGB}{255,165,0}
\newcommand{\QG}[1]{{\color{QGcolor}#1}}
\newcommand{\QR}[1]{{\color{QRcolor}#1}}
\newcommand{\QO}[1]{{\color{Ocolor}#1}}

\def\BibTeX{{\rm B\kern-.05em{\sc i\kern-.025em b}\kern-.08em
    T\kern-.1667em\lower.7ex\hbox{E}\kern-.125emX}}
\begin{document}
\title{Instruction-Free Tuning of Large Vision Language Models for Medical Instruction Following}
\author{Myeongkyun~Kang\orcidlink{0000-0002-9165-870X},
Soopil~Kim\orcidlink{0000-0001-8937-6263},
Xiaoxiao~Li\orcidlink{0000-0002-8833-0244},
and~Sang~Hyun~Park\orcidlink{0000-0001-7476-1046},
\thanks{\textit{Corresponding author: Sang Hyun Park.}}
\thanks{M. Kang and X. Li are with the Department of Electrical and Computer Engineering, The University of British Columbia, Vancouver, BC V6T 1Z4, Canada, and also with the Vector Institute, Toronto, ON M5G 0C6, Canada.}
\thanks{S. Kim is with the Division of Intelligent Robot, Daegu Gyeongbuk Institute of Science and Technology (DGIST), Daegu 42988, Republic of Korea.}
\thanks{S. H. Park is with the Department of Computer Science and Engineering, Pohang University of Science and Technology (POSTECH), Pohang 37673, Republic of Korea. (e-mail: sanghyunpark@postech.ac.kr)}
}
\maketitle
\begin{abstract}
Large vision language models (LVLMs) have demonstrated impressive performance across a wide range of tasks. These capabilities largely stem from visual instruction tuning, which fine-tunes models on datasets consisting of curated image-instruction-output triplets. However, in the medical domain, constructing large-scale, high-quality instruction datasets is particularly challenging because curation by clinical experts is costly and difficult to scale. To address this issue, we propose an instruction-free tuning approach that reduces reliance on manually curated or generated instructions, leveraging only image-description pairs for fine-tuning. Specifically, we introduce a proxy instruction as a replacement for curated text instructions, which preserves the instruction-following capability of the pre-trained LVLM. By gradually updating the proxy instruction via an exponential moving average, we promote updates to parameters that remain valid during inference. Consequently, the fine-tuned LVLM can flexibly respond to domain-specific instructions, even though explicit instructions are absent during fine-tuning. Additionally, we incorporate a response shuffling strategy to mitigate the model's over-reliance on previous words, facilitating more effective fine-tuning. Our approach achieves state-of-the-art accuracy on multiple-choice visual question answering tasks across SKINCON, WBCAtt, CBIS, and MIMIC-CXR datasets, significantly enhancing the fine-tuning efficiency of LVLMs in medical domains.
\end{abstract}

\begin{figure}[t]
\centering
\begin{subfigure}{0.9\linewidth}
\includegraphics[width=\linewidth]{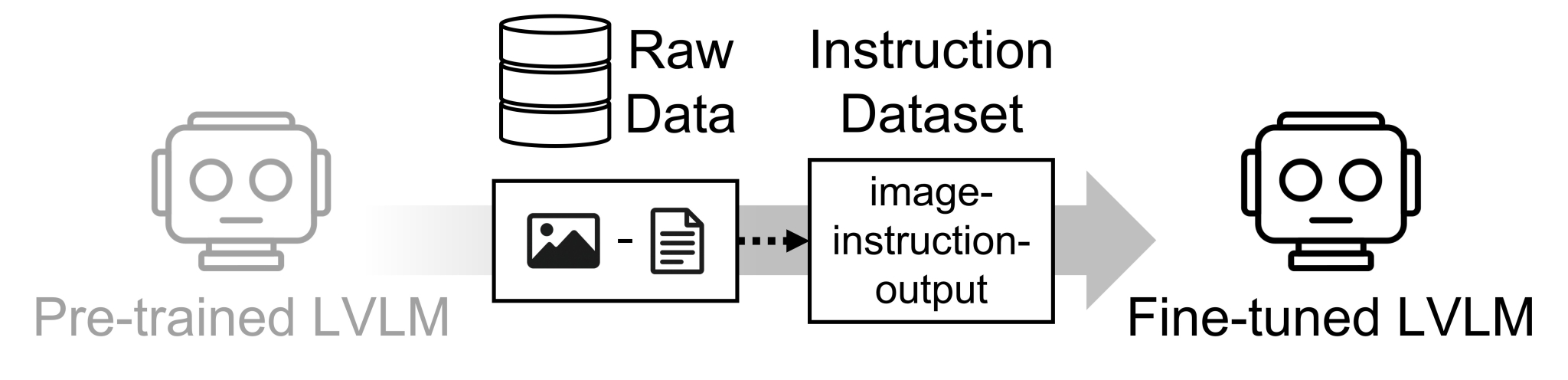}
\caption{Instruction Tuning}
\end{subfigure}
\begin{subfigure}{0.9\linewidth}
\includegraphics[width=\linewidth]{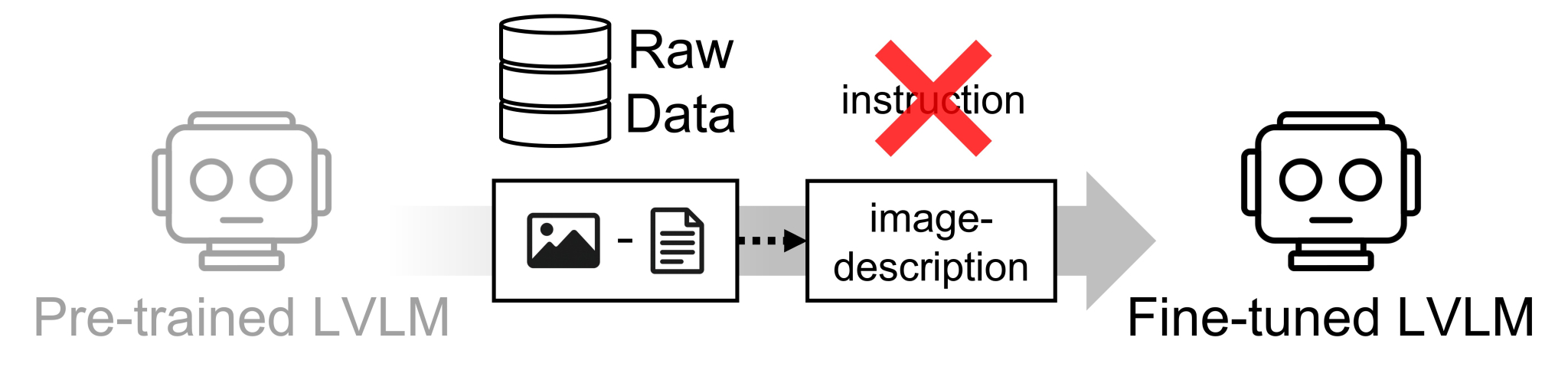}
\caption{Instruction-Free Tuning}
\end{subfigure}
\begin{subfigure}{0.9\linewidth}
\includegraphics[width=\linewidth]{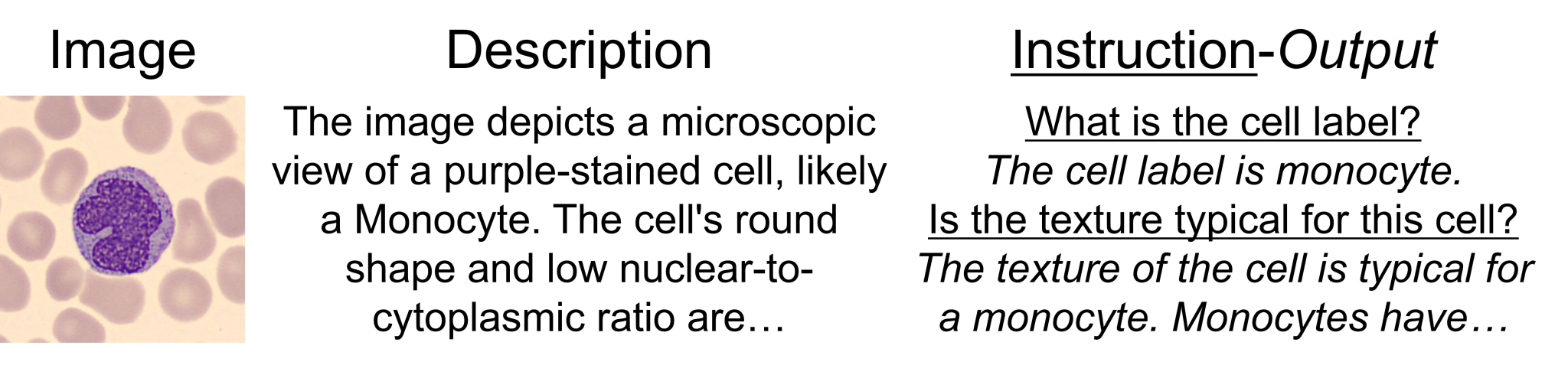}
\caption{Examples of description and instruction-output}
\end{subfigure}
\caption{Conceptual illustrations of (a) instruction tuning, which fine-tunes the model on curated instructions and outputs, and (b) instruction-free tuning, which fine-tunes the model solely on paired textual descriptions (e.g., medical reports). Instruction tuning requires instruction-image-output triplets constructed by humans or LLMs prior to fine-tuning, whereas instruction-free tuning can be performed on image-description pairs without additional steps. Examples of an image, an instruction-output pair for instruction tuning, and a description used for instruction-free tuning are shown in (c).}
\label{fig_concept}
\end{figure}

\section{Introduction}
Large vision language models (LVLMs) have demonstrated strong general-purpose capabilities across a wide range of visual language understanding and generation tasks \cite{liu2023visual}.
This success is largely attributable to large-scale visual instruction tuning, where models are fine-tuned on datasets consisting of image-instruction-output triplets (Fig. \ref{fig_concept}(a)).
Despite its effectiveness in enhancing adherence to diverse human instructions, visual instruction tuning is highly dependent on the quality and domain coverage of the instruction dataset \cite{zhou2023lima}.
As most instruction datasets are built from general-domain data, instruction-tuned general-purpose LVLMs often perform poorly in specialized domains such as medicine \cite{li2023llava}.

Several studies have adapted LVLMs for medical applications by using instruction datasets specifically designed for the medical domain \cite{sellergren2025medgemma,chen2024towards,park2026robust,arsalane2024context}.
Since medical images are typically paired with textual descriptions (e.g., medical reports) rather than in an instruction-output format, these studies leverage language-only large language models (LLMs) to transform image descriptions into conversational image-instruction-output triplets for visual instruction tuning.
However, the frequent use of domain-specific abbreviations and terminology \cite{moon2014sense}, along with the need for expert-level contextual understanding \cite{leaman2015challenges}, limits the quality of instruction datasets constructed using LLMs \cite{fleming2024medalign}.
Conversely, manual curation by clinical experts ensures dataset quality but is costly and difficult to scale \cite{lau2018dataset,fleming2024medalign}.
Consequently, constructing high-quality instruction datasets at scale poses a major bottleneck in adapting LVLMs to the medical domain.

This motivates us to adapt instruction-tuned LVLMs directly from image-description pairs, without constructing instruction datasets.
We revisit fine-tuning approaches that reduce reliance on manually curated or generated instructions (as illustrated in Fig. \ref{fig_concept}(b)).
\emph{Instruction-free tuning} refers to fine-tuning LVLMs using only image-description pairs (e.g., medical reports), without relying on any instructions.
To the best of our knowledge, this is the first work to formalize this task for instruction-tuned LVLMs.
The objective of instruction-free tuning is to adapt the LVLM to the medical domain while preserving its pre-trained instruction-following capability, enabling the fine-tuned model to respond to diverse domain-specific instructions at inference.

Since instruction-tuned LVLMs are trained to respond to textual instructions, a straightforward approach is to assign a fixed instruction (e.g., \emph{``Describe this medical scan''}) to all image-description pairs \cite{liu2023visual,li2023llava,chen2024towards}.
However, paired descriptions (e.g., medical reports) do not perfectly align with the natural response (e.g., \emph{``The image...''}) that would typically follow a conversational instruction, which poses a risk of degrading the instruction-following capability of the pre-trained LVLM.
Specifically, this may lead to \emph{image-dependence}, where the model generates fixed-format responses that rely excessively on the image regardless of the given instruction (see Fig. \ref{fig_qual}) \cite{wang2024two}.
Additionally, this may result in \emph{instruction-dependence}, where the model responds appropriately only to instructions encountered during fine-tuning and fails to generate responses aligned with the fine-tuning data for unseen instructions (see other LVLMs in Table \ref{tab_main}) \cite{kotha2024understanding}.

Another challenge is that medical reports typically begin with a lesion description and conclude with a diagnostic impression \cite{kahn2009toward}, which leads to strong inter-sentence confounding dependencies within the descriptions.
As a result, diagnostic conclusions can often be inferred from the preceding text alone (e.g., findings), leading models to overlook visual features when generating the later part of the response and exacerbating hallucinations \cite{favero2024multi,kim2026rethinking}.

In response to the first challenge, we introduce a \emph{proxy instruction} that is optimized to align with the set of image-description pairs in the fine-tuning dataset.
Unlike conventional fine-tuning, which uses text instructions and optimizes model parameters, our method jointly updates both learnable proxy instructions and model parameters.
This enables the LVLM to adapt to the medical domain without degrading its instruction-following capability through a well-aligned instruction.
Furthermore, we update the proxy instruction more slowly than the vision encoder through an exponential moving average, i.e., a \emph{gradual proxy update}.
Since the proxy instruction is discarded at inference, this ensures that the adaptation is learned mainly by the retained vision encoder rather than by the proxy instruction.
This differs from existing prompt-learning methods for text-only LLMs, whose learnable prompts are fixed during fine-tuning \cite{wang2024two} or retained at inference \cite{li2021prefix}.

To address the second challenge, we propose a strategy, \emph{response shuffling}, which randomly shuffles the sentence order of the ground truth description, thereby preventing the model from relying too heavily on previous words.
Notably, employing a proxy instruction mitigates the negative effects of shuffling output order during fine-tuning, enabling a simple strategy to substantially enhance medical visual question answering (VQA) performance.

In summary, the contributions of this paper are as follows:
\begin{itemize}
\item We introduce and formalize \emph{instruction-free tuning}, a novel paradigm for fine-tuning LVLMs that eliminates the need for manually curated or generated instructions, addressing the major bottleneck in adapting LVLMs to the medical domain.
\item We propose a \emph{proxy instruction} with a gradual proxy update, designed to preserve the instruction-following capability of the pre-trained LVLM during fine-tuning, while promoting updates for parameters that remain valid during inference.
Additionally, we propose a \emph{response shuffling} strategy to mitigate the model's over-reliance on previous words, facilitating more effective fine-tuning of medical LVLMs.
\item We demonstrate that our fine-tuned model achieves state-of-the-art accuracy in multiple-choice VQA evaluations on the SKINCON, WBCAtt, CBIS, and MIMIC-CXR datasets.
\end{itemize}
\section{Related Works}
\subsection{Instruction Tuning}
Instruction tuning fine-tunes a model using a dataset consisting of instruction-input-output triplets.
In particular, the instruction specifies the task (i.e., a question), the input provides supplementary context (e.g., an image), and the output is the expected response based on the instruction and input.
As one of the early studies, Natural Instructions \cite{mishra2022cross} fine-tuned a model with an instruction dataset constructed by integrating 61 datasets.
Recently, an alternative approach that constructs instruction datasets using LLMs was proposed; Alpaca \cite{taori2023alpaca} efficiently fine-tuned the LLaMA \cite{touvron2023llama} using 52K instruction-output pairs generated by an instruction-tuned GPT-3.5 \cite{achiam2023gpt}.
Moreover, Self-Instruct \cite{wang2023self} proposed a method for enabling an LLM to follow diverse instructions by iteratively generating and fine-tuning on synthetic instruction-output data with minimal human intervention.
However, applying these methods to visual data is challenging because image generation in pixel space is inherently more difficult than text generation. Therefore, prior work has focused on constructing instruction datasets from paired image-text data (e.g., LLaVA \cite{liu2023visual}), rather than relying on synthetic data generation. However, the quality of LLM-generated instruction data is bounded by the domain knowledge of the LLM, limiting its benefit in specialized domains.
In the medical domain, for example, even GPT-4 \cite{achiam2023gpt} failed on 35\% of clinician-written instructions over electronic health records, whereas expert annotation ensures quality but is costly and difficult to scale \cite{fleming2024medalign}.

\subsection{Visual Instruction Tuning}
Building on the success of instruction tuning, visual instruction tuning with LLMs was introduced for vision tasks.
LLaVA \cite{liu2023visual} integrated CLIP's vision encoder \cite{radford2021learning} with LLaMA \cite{touvron2023llama}, and was fine-tuned on an instruction dataset generated by language-only GPT-4 \cite{achiam2023gpt}.
Recently, LLaMA-3.2-Vision \cite{dubey2024llama} extended LLaMA 3.1 by adapting cross-attention layers to integrate the image modality with language.
Additionally, alternative LLMs have been integrated into the vision domain, such as Qwen2.5-VL \cite{bai2025qwen2}, which utilizes patch grouping and window attention to efficiently process large numbers of visual tokens.

These successes in LVLMs extended to medical domains.
LLaVA-Med \cite{li2023llava} was fine-tuned on an instruction dataset generated by GPT-4 \cite{achiam2023gpt}, based on biomedical image-caption pairs from PubMed Central (PMC).
Recently, PubMedVision \cite{chen2024towards} and MedGemma \cite{sellergren2025medgemma} have been fine-tuned on carefully curated medical datasets from PMC and other publicly available sources, significantly enhancing their medical capabilities.
Lingshu \cite{xu2026lingshu} further synthesized large-scale medical multimodal data, where medical professionals reviewed only small pilot batches of synthesized captions.
However, clinician verification of all instruction data does not scale; accordingly, clinician-constructed datasets remain small \cite{lau2018dataset,fleming2024medalign}.

\subsection{Fine-Tuning without Text Instructions}
Prepending learnable continuous vectors (i.e., learnable prompts) to the input has been explored for adapting language models.
Prefix-Tuning \cite{li2021prefix} optimizes continuous prompts prepended to the input while keeping the language model frozen, and the learned prompts are used at inference.
For text-only LLMs, ProMoT \cite{wang2024two} trains a learnable prompt and then fine-tunes the model with the prompt fixed.
The prompt thereby absorbs the task-specific format, and it is kept for the fine-tuned task but discarded for tasks with different formats.
Unlike these methods, our proxy instruction is gradually updated together with the vision encoder rather than fixed, and is always discarded at inference.
One study \cite{liu2025vift} attempted instruction-free tuning by using language-only instruction-output pairs and image-caption pairs with a fixed set of instructions.
However, it still relies on text instructions during LVLM fine-tuning.
To our knowledge, this is the first approach to enable fine-tuning of instruction-tuned LVLMs without using any text instructions.

\section{Method}

\subsection{Overview}
\begin{figure*}[t]
    \centering
    \includegraphics[width=0.9\textwidth]{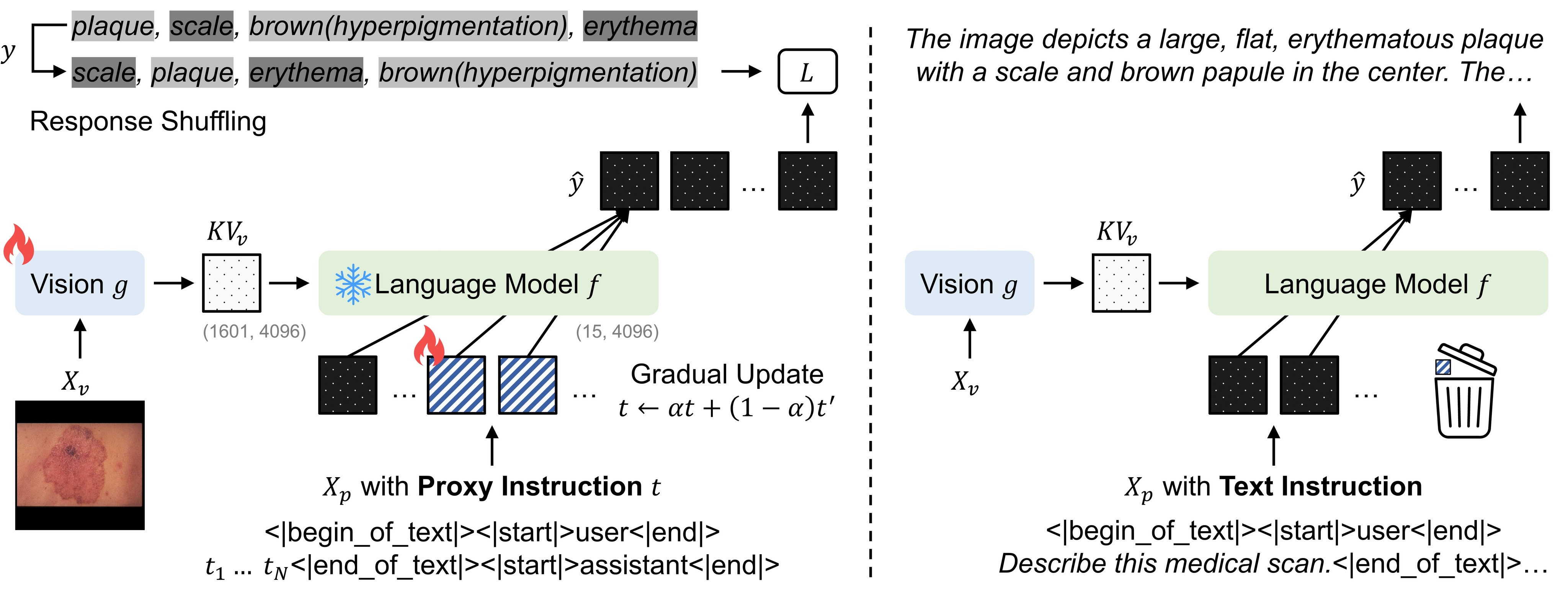}
    \caption{An illustration of our instruction-free tuning framework. The vision encoder $g$ extracts key-value matrices ${KV}_v$ from an image $X_v$, which are then integrated into a prompt $X_p$ for the language model $f$ to generate the response $\hat{y}$. For instruction-free tuning, the text instruction in $X_p$ is replaced with the proxy instruction $t$. During supervised fine-tuning, $g$ is updated with autoregressive loss $L$ between the response $\hat{y}$ and the ground truth $y$. In parallel, starting from the warm-up stage, the proxy instruction $t$ is gradually updated toward its gradient-updated value $t'$ with the EMA decay rate $\alpha$. During inference, $t$ is discarded, and a conversational text instruction (e.g., \emph{``Describe...''}) is used to generate a natural language response (e.g., \emph{``The image depicts...''}).}
    \label{fig_method}
\end{figure*}

Our framework is illustrated in Fig. \ref{fig_method}.
It comprises a vision encoder $g$ (a vision transformer \cite{dosovitskiy2020image} followed by projection layers) and a language model $f$.
Given an image $X_v$, a prompt $X_p$ containing an instruction (e.g., a question), and the ground truth output $y$ (e.g., an answer or description), the image $X_v$ is fed into $g$ to extract key-value matrices ${KV}_v$ for generating responses conditioned on the image and the instruction.
Specifically, $f$ employs a cross-attention layer \cite{dubey2024llama, alayrac2022flamingo} to integrate the query matrix derived from $X_p$ with ${KV}_v$ to generate the response $\hat{y}$.
During fine-tuning, we compute an autoregressive loss $L$ between $y$ and $\hat{y}$ to update the vision encoder $g$, while keeping the parameters of $f$ frozen, following \cite{torchtune}.
The following section provides details of supervised fine-tuning for \emph{instruction tuning} and \emph{instruction-free tuning}.

\subsection{Instruction Tuning}
We start by formulating \emph{instruction tuning}.
Instruction tuning fine-tunes the model on an image $X_v$, a prompt $X_p$ containing a text instruction (i.e., a question), and a ground-truth output $y$ (i.e., an answer).
Formally, supervised fine-tuning for instruction tuning is defined as follows:
\begin{equation}\label{eq_loss_pre}
L(g) = - \sum_{i \in \mathcal{I}_Y} \log f\bigl(y_i \mid X_v, X_p, y_{<i}; g\bigr),
\end{equation}
where $\mathcal{I}_Y$ denotes the set of indices corresponding to the output sequence $y$, and $y_{<i}$ represents all previously generated tokens.
However, constructing high-quality instruction datasets consisting of carefully curated instruction-output pairs corresponding to medical images is challenging and often incurs substantial costs.
As a result, prior studies \cite{chen2024towards,li2023llava} leverage inherent properties of medical data as text instructions (e.g., \emph{``Describe this medical scan''}) to construct instruction datasets.
In practice, paired descriptions (e.g., medical reports) do not perfectly align with the natural conversational responses expected from text instructions.
This misalignment between instruction-description pairs in fine-tuning and pre-training datasets risks degrading the instruction-following capability of pre-trained LVLMs.
To address these issues, we introduce \emph{instruction-free tuning}, which reduces reliance on curated instructions by employing only image-description pairs during fine-tuning.

\subsection{Instruction-Free Tuning}
In instruction-free tuning, the model is fine-tuned using an image $X_v$, a prompt $X_p$ containing a \emph{proxy instruction}, and a ground-truth description $y$.
Unlike instruction tuning, which requires curated text instructions during fine-tuning, our approach employs a proxy instruction that is optimized to align with image-description pairs.
%
Let the proxy instruction $t=\{t_1,\ldots,t_N\}$ be defined as a set of $N$ continuous vectors, each with the same dimensionality as the word embeddings.
During fine-tuning, the proxy instruction is placed at the position of the text instruction (e.g., a question) in the prompt $X_p$, leaving the remaining tokens unchanged, and at inference, it is replaced with a text instruction.
Since $f$ is differentiable, gradients can be propagated to update $t$, enabling the model to derive a well-aligned instruction from the dataset.
To optimize $t$, we compute an autoregressive loss $L$ between $y$ and $\hat{y}$, as defined in Eq.~\ref{eq_loss_pre}, with the inclusion of $t$.
Formally, supervised fine-tuning for instruction-free tuning is defined as:
\begin{equation}\label{eq_loss}
L(g,t) = - \sum_{i \in \mathcal{I}_Y} \log f\bigl(y_i \mid X_v, X_p, y_{<i}; g,t\bigr).
\end{equation}
For simplicity, supplementary tokens in $X_p$, such as \texttt{<|eot\_id|>}, are omitted.
The proxy instruction is optimized to align with image-description pairs, thereby mitigating the negative impact on the pre-trained instruction-following capability of LVLMs during fine-tuning \cite{wang2024two}.
This approach is particularly suitable for medical domains, since it is infeasible to fully provide all possible instructions during fine-tuning.
Notably, the proxy instruction differs from that used in Prefix-Tuning \cite{li2021prefix}. In Prefix-Tuning, the learnable vectors are used during inference, whereas in our method they serve as a proxy during fine-tuning and are discarded at inference.

Instruction-free tuning proceeds in two stages, i.e., a warm-up stage and a fine-tuning stage with a gradual proxy update, as summarized in Algorithm \ref{algo}.

\subsubsection{Warm-up Stage}
We randomly initialize $t$ and optimize only $t$ via Eq.~\ref{eq_loss} while keeping $g$ frozen.
This aligns $t$ with the image-description pairs before $g$ is updated.
Because $t$ is updated only slowly under the gradual proxy update, this warm-up ensures that $g$ is fine-tuned with an aligned proxy instruction rather than its random initialization.

\subsubsection{Gradual Proxy Update}
After warm-up, we jointly fine-tune $g$ and $t$.
Although the optimized instruction $t$ is aligned with its corresponding descriptions, it is replaced with a conversational text instruction at inference.
Hence, if $t$ absorbs part of the adaptation during fine-tuning, this adaptation is discarded together with $t$, degrading the performance of the fine-tuned LVLM.
One way to mitigate this issue is to fix the proxy instruction after warm-up \cite{wang2024two}; however, a fixed proxy instruction cannot follow the changes in $g$ during fine-tuning.
Instead, we update the proxy instruction more slowly than $g$ via an exponential moving average (EMA), so that it serves as a stable anchor.
Formally, at each iteration, we compute $t' \leftarrow t - \eta \nabla_{t} L(g,t)$ and then update $t$ as follows:
\begin{equation}\label{eq_gpu}
t \leftarrow \alpha t + (1 - \alpha) t',
\end{equation}
where $\alpha\in [0, 1)$ denotes the EMA decay rate and $\eta$ denotes the learning rate.
To show how $\alpha$ controls the update speed of $t$, substituting $t'$ into Eq.~\ref{eq_gpu} yields
\begin{equation}\label{eq_equiv}
t \leftarrow t - (1 - \alpha)\, \eta\, \nabla_{t} L(g,t).
\end{equation}
This indicates that the gradual proxy update is equivalent to updating the proxy instruction with a learning rate scaled by $(1-\alpha)$.
Importantly, this reduced learning rate applies only to the proxy instruction, which is discarded at inference, whereas $g$, which is retained, is updated with the learning rate $\eta$.
Consequently, the adaptation to the fine-tuning dataset is driven mainly by $g$.

\begin{algorithm}[t]
\small
\caption{Instruction-free tuning process.}
\label{algo}
\begin{algorithmic}[1]
\STATE \textbf{Input:} Vision encoder $g$, language model $f$, ground truth description $y$, learning rate $\eta$, EMA decay rate $\alpha$
\STATE Randomly initialize $t$
\WHILE{not warmed-up}
    \STATE $y \leftarrow \texttt{RS}(y)$ \quad \texttt{// Response Shuffling.~See Eq.\ref{eq_rs}}
    \STATE \textbf{Compute} $\mathcal{L} \leftarrow L(g,t)$ \quad \texttt{// See Eq.\ref{eq_loss}}
    \STATE $t \leftarrow t - \eta \nabla_t \mathcal{L}$
\ENDWHILE
\WHILE{not converged}
    \STATE $y \leftarrow \texttt{RS}(y)$
    \STATE \textbf{Compute} $\mathcal{L} \leftarrow L(g,t)$
    \STATE $g \leftarrow g - \eta \nabla_g \mathcal{L};\quad t' \leftarrow t - \eta \nabla_{t} \mathcal{L}$
    \STATE $t \leftarrow \alpha t + (1 - \alpha) t'$
\ENDWHILE
\end{algorithmic}
\end{algorithm}

\subsection{Response Shuffling}
In medical domains, the model tends to over-rely on previous words $y_{<i}$ during supervised fine-tuning.
Specifically, medical reports typically begin with lesion descriptions and conclude with diagnostic impressions.
When the order is fixed, later text (e.g., the diagnosis) can be inferred from the preceding text alone.
To address this issue, we randomly shuffle $y$ during fine-tuning.
Formally,
\begin{equation}\label{eq_rs}
\texttt{RS}(y) = \texttt{Join}(\texttt{Shuffle}(\texttt{Split}(y))),
\end{equation}
where $\texttt{Split}$ segments $y$ by a separator (e.g., ``,''), $\texttt{Shuffle}$ randomly permutes the segments, and $\texttt{Join}$ reconstructs the permuted $y$ using the original separator.
The segments are reshuffled whenever a sample is loaded.
Since the segment order varies across iterations, the next segment can no longer be inferred from the preceding segments, which reduces over-reliance on previous words.
Consequently, response shuffling improves the reliability of individual attribute predictions rather than merely the overall accuracy, particularly for key attributes (e.g., see label in Fig.~\ref{fig_radar}(a)).
Notably, the goal of our fine-tuning is not to train the model to generate high-quality medical reports, but rather to adapt the model to a domain-specific dataset while preserving the pre-trained instruction-following capability.
Therefore, the negative effects of shuffling the output order during fine-tuning are mitigated (see Fig. \ref{fig_qual}).
\section{Experiments}

\begin{figure}[t]
    \centering
    \includegraphics[width=0.85\linewidth]{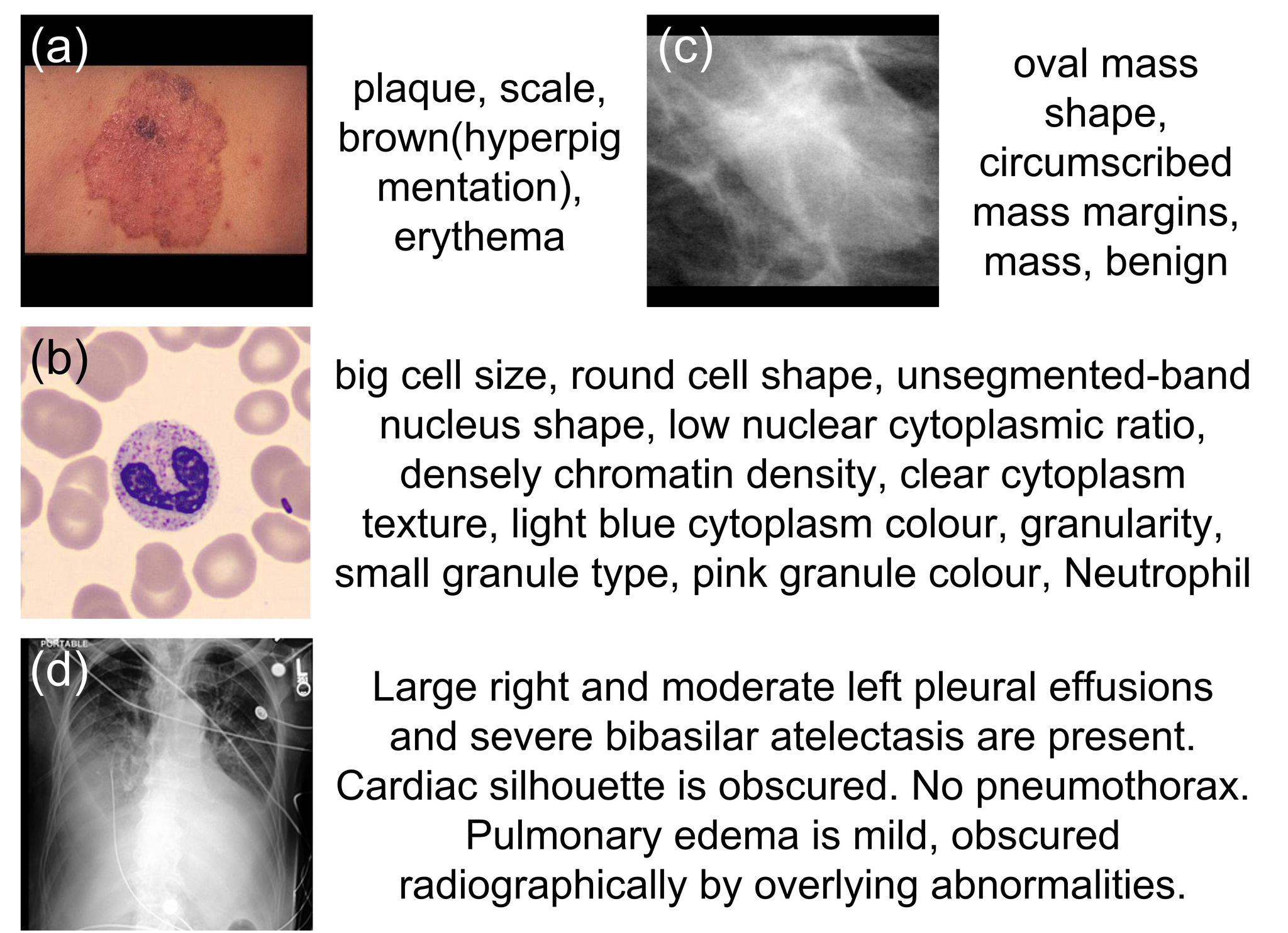}
    \caption{Examples of the medical report for the (a) SKINCON, (b) WBCAtt, (c) CBIS, and (d) MIMIC-CXR datasets.}
    \label{fig_dataset}
\end{figure}

\subsection{Datasets}
We conducted experiments on four richly annotated medical datasets, as shown in the Fig. \ref{fig_dataset}.

\begin{itemize}
\item \textbf{SKINCON} \cite{daneshjou2022skincon} is a dermatology dataset that contains 3230 images from the Fitzpatrick17k \cite{groh2021evaluating}. Each image was manually annotated by dermatologists using 48 clinical attributes, such as \emph{plaque}. The dataset was split into 80\% for training, 10\% for validation, and 10\% for testing.
\item \textbf{WBCAtt} \cite{tsutsui2023wbcatt} is a dataset consisting of 10298 microscopic images of white blood cells, such as \emph{neutrophils}. Each image was annotated with 11 attributes, such as \emph{cell size}, based on guidelines defined by pathologists. The dataset was split into 6179 images for training, 1030 for validation, and 3099 for testing. 
\item \textbf{CBIS} \cite{lee2017curated} is a digitized mammography dataset consisting of 892 mass cases and 753 calcification cases. Each mammogram was annotated with \emph{pathology}, \emph{abnormality types}, and BI-RADS descriptors (e.g., \emph{mass shape}). This dataset provides cropped images of lesions extracted through ROI segmentation. For our experiments, we used cropped images along with their corresponding attribute annotations. The dataset was split into 90\% for training and 10\% for validation, and the official test split was used for testing.
\item \textbf{MIMIC-CXR} \cite{johnson2019mimic} is a chest X-ray dataset that contains 377110 images corresponding to 227835 radiographic studies.
Each study is paired with a free-text radiology report obtained from the Beth Israel Deaconess Medical Center’s electronic health record system for the period 2011-2016, and we used version 2.1.0 for our experiments.
To extract image-relevant content from free-text reports, we used the findings and impression sections of LLM-annotated MIMIC-CXR reports from \cite{zambrano2025clinically}.
To select frontal chest X-rays, we employed CheXmask \cite{gaggion2024chexmask} to filter out non-chest X-ray images with a Reverse Classification Accuracy (RCA) below 0.7.
We followed the official MIMIC-CXR split for training, validation, and testing.
\end{itemize}

\begin{figure}[t]
    \centering
    \includegraphics[width=1.0\linewidth]{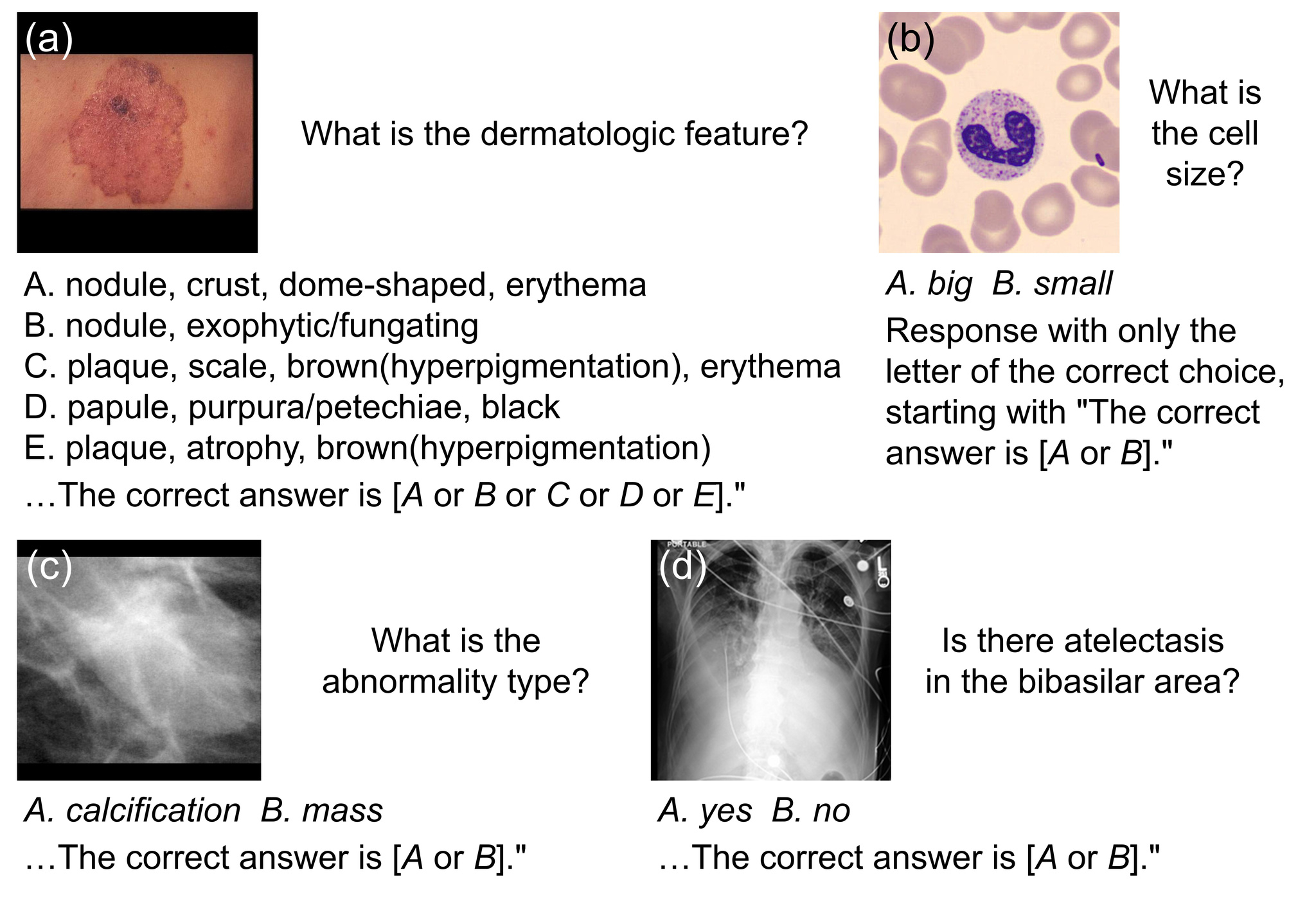}
    \caption{Examples of multiple-choice VQA for the (a) SKINCON, (b) WBCAtt, (c) CBIS, and (d) MIMIC-CXR datasets.}
    \label{fig_vqa}
\end{figure}

\subsection{Data Preparation for Instruction-Free Evaluation}
To evaluate the fine-tuned LVLMs in instruction-free tuning scenarios, we converted the training split data into a medical report (see Fig. \ref{fig_dataset}), while the validation and test split data into a multiple-choice VQA (see Fig. \ref{fig_vqa}).
When converting to medical reports, attribute annotations were formatted into plain text.
In the \textbf{SKINCON} dataset, a medical report was formatted by listing the names of the items marked as \emph{present} among the 48 attributes, such as \emph{``plaque, scale.''}
In the \textbf{WBCAtt} dataset, a medical report was formatted by listing 8 attributes in the form of \emph{\{value\} \{name\}} (e.g., \emph{``big cell size''}).
Meanwhile, yes/no attributes, such as \emph{granularity}, were added using the \emph{\{name\}} when present, whereas \emph{label} attributes were added using the \emph{\{value\}}.
In the \textbf{CBIS} dataset, a medical report was formatted by listing the \emph{\{value\}} of \emph{abnormality} and \emph{pathology} attributes. The remaining attributes were added in the form of \emph{\{value\} \{name\}}.
For the \textbf{MIMIC-CXR} dataset, the preprocessed reports from \cite{zambrano2025clinically} were used without further processing.

When converting to multiple-choice VQA, attribute annotations were formatted into question-answer pairs, usually with five options.
The question was constructed using the attribute name in the form \emph{``What is the \{name\}?''}, whereas for the SKINCON dataset, the fixed question \emph{``What is the dermatologic feature?''} was used.
For multiple-choice options, the annotated attribute value was set as the correct answer, while randomly sampled non-overlapping values from the same attribute served as the incorrect options.
Attributes with fewer than five options (e.g., yes/no) were converted into a limited choice set (e.g., two options).
For \textbf{MIMIC-CXR} evaluation, we used the Medical-CXR-VQA dataset \cite{hu2024interpretable}.
This dataset consists of VQA pairs for chest X-ray interpretation built from the MIMIC-CXR dataset, covering questions on abnormalities, presence, view, location, level, and type.
Since we used only frontal chest X-rays for our experiments, we excluded view-related question types.
For consistency with the other datasets, we added incorrect options to convert the dataset into a multiple-choice VQA format with five options.
Incorrect options were set to different values within the same question type, while yes/no questions were limited to two options.
For abnormality-type questions with multiple disease answers, we conducted VQA separately for each disease, setting each abnormality as a separate correct choice.
To promote reproducibility and fair comparison, we release the full instruction-free tuning benchmark for the SKINCON, WBCAtt, and CBIS datasets, including the preprocessing pipelines for the medical reports and the multiple-choice VQA evaluation protocols.

\subsection{Implementation Details}
The proposed method was developed on top of LLaMA-3.2-11B-Vision-Instruct \cite{dubey2024llama}, and fine-tuning was performed using the torchtune framework \cite{torchtune}.
For models with proxy instructions, the number of instruction vectors $N$ was set to 8, and the EMA decay rate $\alpha$ was set to 0.999.
The warm-up stage was conducted for one epoch, while the fine-tuning stage was conducted for five epochs.
We mainly followed the default settings in torchtune for supervised fine-tuning, such as a batch size of 2 and the AdamW optimizer.
We use an input resolution of 560$\times$560 for fine-tuning and evaluation.
For all datasets, we specified the response style at the end of each question (e.g., \emph{``Response with$\dots$''}; see Fig. \ref{fig_vqa}) and extracted the predicted choice (e.g., A) by parsing the model’s output.
We evaluated multiple-choice VQA using accuracy by comparing the selected choices with the correct answers.
We conducted each experiment three times with different random seeds for the main results and report the average accuracy.
As demographic annotations were not consistently available across all datasets, these variables were not incorporated into the analysis.
\begin{table*}[t]
\centering
\caption{Multiple-choice VQA accuracy on the SKINCON, WBCAtt, and CBIS datasets. We compared our method (InstFree \w RS) with BLIP-2 \cite{li2023blip}, Qwen2.5-VL-3B \cite{bai2025qwen2}, LLaMA-3.2-11B-Vision \cite{dubey2024llama}, LLaVA-Med \cite{li2023llava}, MedGemma-4B \cite{sellergren2025medgemma}, PubMedVision-7B \cite{chen2024towards}, and Lingshu-7B \cite{xu2026lingshu}. FT denotes fine-tuning, and RS denotes response shuffling. \textbf{Bold} indicates the highest accuracy.}
\label{tab_main}
\begin{tabular}{l|c|lll|l}
\hline
Model & Medical & SKINCON & WBCAtt & CBIS & Avg. \\
\hline
BLIP-2 (\wo FT) & \multirow{3}{*}{$\times$} & 8.7 & 35.2 & 4.6 & 16.1 \\
Qwen2.5-VL-3B (\wo FT) & & 37.7 & 32.8 & 36.0 & 35.5 \\
LLaMA-3.2-11B-Vision (\wo FT) & & 39.9 & 34.6 & 43.0 & 39.1 \\
\hline
LLaVA-Med (\wo FT) & \multirow{4}{*}{$\checkmark$} & 1.6 & 32.8 & 36.7 & 23.7 \\
MedGemma-4B (\wo FT) & & 14.3 & 37.3 & 39.2 & 30.2 \\
PubMedVision-7B (\wo FT) & & 47.0 & 36.3 & 40.6 & 41.3 \\
Lingshu-7B (\wo FT) & & 54.2 & 51.6 & 55.8 & 53.9 \\
\hline
Qwen2.5-VL-3B (FT) & \multirow{4}{*}{--} & 42.0 \SD{0.42} & 39.4 \SD{0.19} & 39.3 \SD{0.12} & 40.2 \\
LLaMA-3.2-11B-Vision (FT) & & 59.4 \SD{3.85} & 61.7 \SD{3.72} & 63.9 \SD{0.3} & 61.6 \\
MedGemma-4B (FT) & & 8.9 \SD{1.76} & 45.9 \SD{1.18} & 42.5 \SD{2.23} & 32.4 \\
InstFree \w RS & & \textbf{75.2} \SD{0.9} & \textbf{65.2} \SD{1.39} & \textbf{68.3} \SD{0.37} & \textbf{69.5} \\
\hline
\end{tabular}
\end{table*}

\begin{figure*}[t]
    \centering
    \begin{subfigure}{0.3\linewidth}
        \centering
        \includegraphics[width=1.0\linewidth]{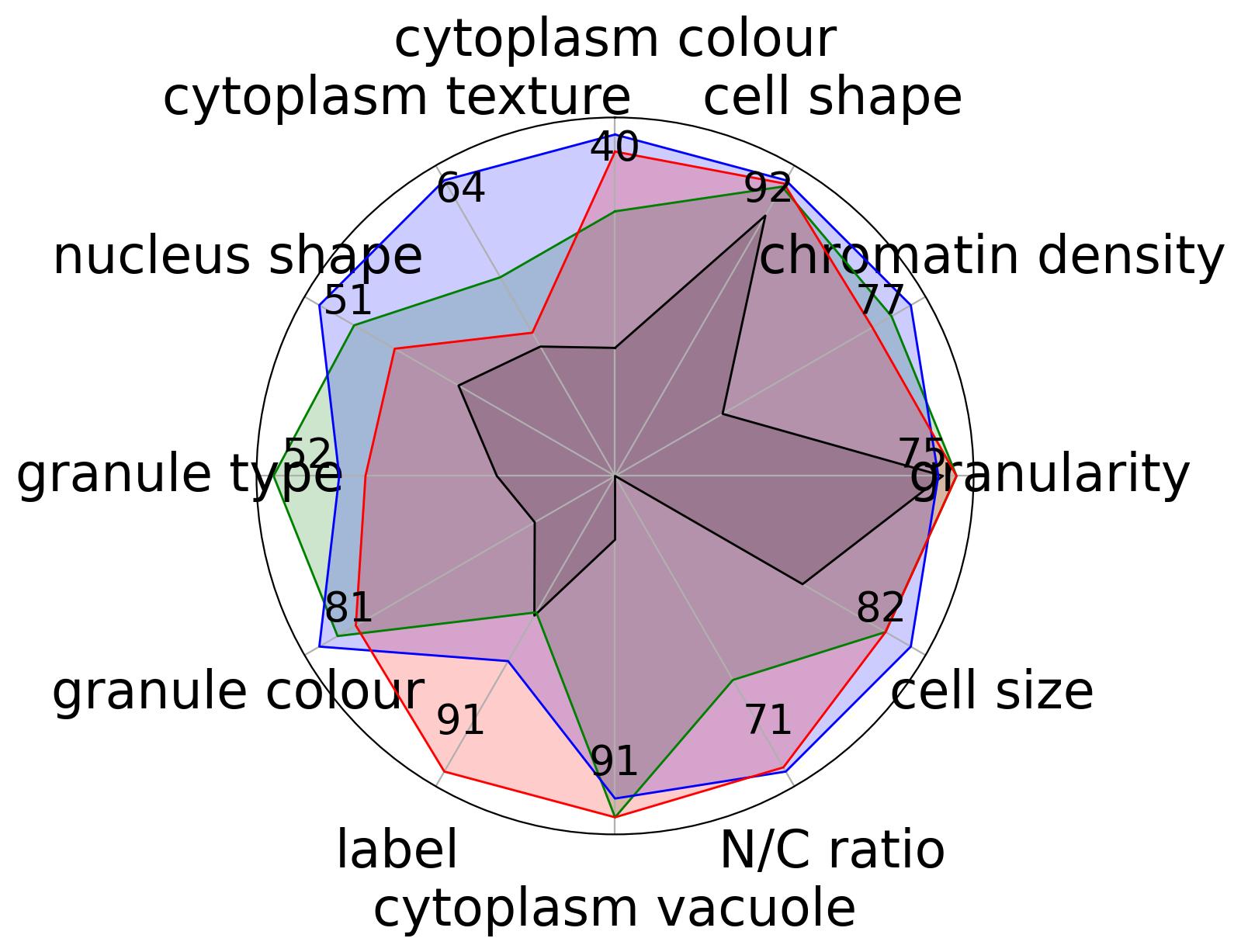}
        \caption{WBCAtt}
    \end{subfigure}
    \begin{subfigure}{0.3\linewidth}
        \centering
        \includegraphics[width=1.0\linewidth]{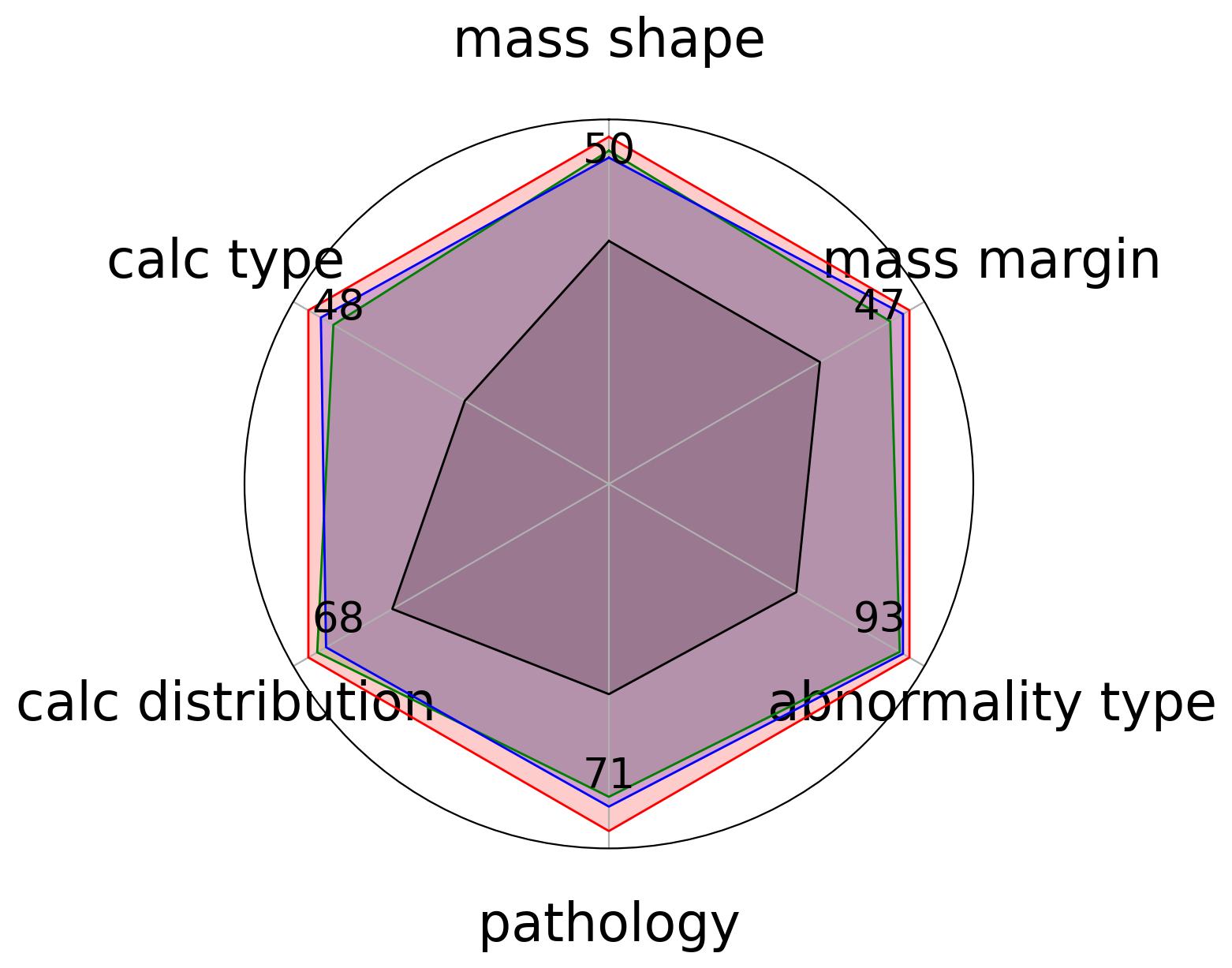}
        \caption{CBIS}
    \end{subfigure}
    \begin{subfigure}{0.3\linewidth}
        \centering
        \includegraphics[width=1.0\linewidth]{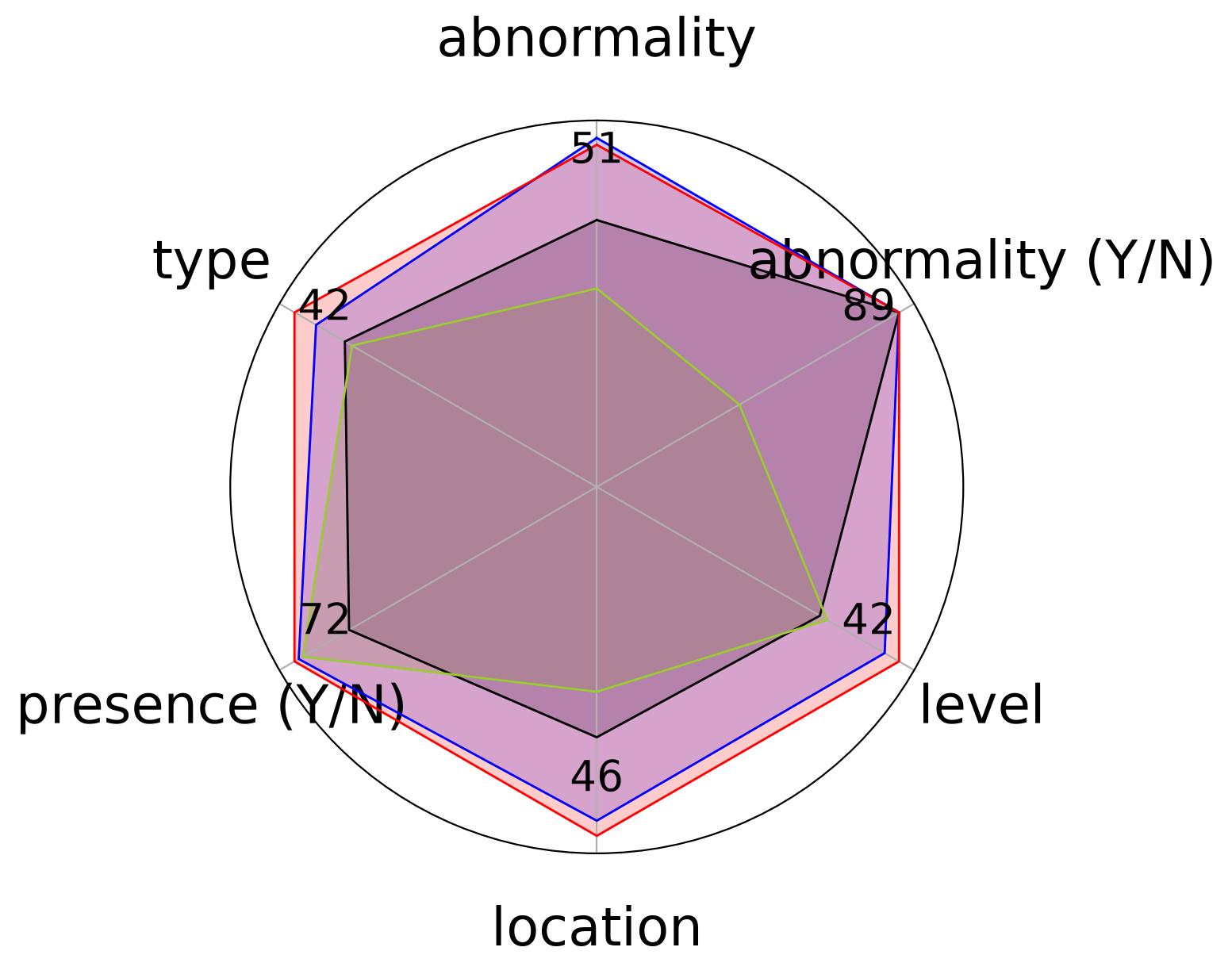}
        \caption{MIMIC-CXR}
    \end{subfigure}
    \caption{Radar charts of the accuracy for each attribute (or question type) across the (a) WBCAtt, (b) CBIS, and (c) MIMIC-CXR datasets. LLaMA-3.2-11B-Vision (\wo FT) is shown in black, LLaMA-3.2-11B-Vision (FT) in green, InstFree in blue, InstFree \w RS in red, and MedGemma-4B in yellow-green.}
    \label{fig_radar}
\end{figure*}

\newpage
\section{Results}
\subsection{Multiple-choice VQA}
\subsubsection{Main Results}
We first compared our method against other LVLMs without fine-tuning (\wo FT) on the SKINCON, WBCAtt, and CBIS datasets, including general LVLMs such as Qwen2.5-VL-3B-Instruct \cite{bai2025qwen2} and LLaMA-3.2-11B-Vision-Instruct \cite{dubey2024llama}, as well as medical LVLMs such as LLaVA-Med \cite{li2023llava}, MedGemma-4B-it \cite{sellergren2025medgemma}, PubMedVision-7B-Qwen2.5VL \cite{chen2024towards}, and Lingshu-7B \cite{xu2026lingshu}.
We then compared our method with fine-tuned (FT) Qwen2.5-VL-3B, LLaMA-3.2-11B-Vision, and MedGemma-4B using a text instruction (i.e., \emph{``Describe this medical scan''}).
Fine-tuning of Qwen2.5-VL-3B and MedGemma-4B followed the settings described in their original papers.
Additionally, we compared our method with BLIP-2 \cite{li2023blip}, which is trained on image-text pairs and is capable of performing VQA tasks. Note that the language model in BLIP-2 is not instruction-tuned and thus has limited instruction-following ability compared to the other LVLMs in our comparison.
We refer our method as \textbf{InstFree}, which fine-tuned the model using the proxy instruction with gradual proxy update, while \textbf{InstFree \w RS} fine-tuned it with both the gradual proxy update and response shuffling.
Since the WBCAtt and CBIS datasets provided attribute names, we reported the accuracy for each attribute separately.

Table \ref{tab_main} shows the multiple-choice VQA accuracy on the SKINCON, WBCAtt, and CBIS datasets.
Overall, InstFree \w RS outperforms all compared LVLMs on every dataset, achieving an average accuracy of 69.5\%.
This demonstrates the effectiveness of the proposed method, which combines the proxy instruction with gradual proxy update and response shuffling.
Among the LVLMs without fine-tuning (\wo FT), medical LVLMs such as PubMedVision-7B and Lingshu-7B outperform general LVLMs on average.
However, MedGemma-4B on SKINCON and LLaVA-Med show lower accuracy than general LVLMs, likely due to limited training data coverage and limited instruction-following capability, respectively.
With fine-tuning, all LVLMs achieve higher average accuracy than their \wo FT counterparts.
These results highlight that fine-tuning LVLMs is crucial for downstream tasks in the medical domain, as it consistently yields performance improvements, even when naively incorporating a text instruction during fine-tuning.

The accuracy varies across fine-tuned LVLMs, with Qwen2.5-VL-3B (FT) and MedGemma-4B (FT) performing significantly worse than LLaMA-3.2-11B-Vision (FT).
This discrepancy is likely attributable to their strong dependence on instructions, i.e., \emph{instruction-dependence}, rather than to the limited model capacity of Qwen2.5-VL-3B and MedGemma-4B.
Specifically, the fine-tuned model generates correct responses to certain instructions (i.e., \emph{``Describe this medical scan''}), but fails to respond appropriately to other types of instructions, such as \emph{``What is the cell shape?'',} instead responding based on its inherent pre-trained knowledge \cite{kotha2024understanding}.
Notably, even though MedGemma-4B has been fine-tuned, it still shows low accuracy on the SKINCON dataset.
This is likely due to the absence of explicit attribute names in the medical reports and the limited instruction-following capability of the pre-trained LVLM, which further exacerbates the \emph{instruction-dependence} of the fine-tuned model.
In summary, the limited ability of fine-tuned Qwen2.5-VL-3B and MedGemma-4B to generalize across diverse instructions leads to a significant degradation in multiple-choice VQA performance.

\noindent\textbf{Scope of instruction-following.} Instruction-free tuning aims to preserve the ability to follow domain-specific instructions relevant to the fine-tuning dataset, rather than general-purpose instructions.
This is because fine-tuning mainly affects inputs close to the fine-tuning distribution, whereas pre-trained capabilities are largely retained on distant ones \cite{kotha2024understanding}.
This tendency is also observed for instruction (c) in Fig.~\ref{fig_qual}.

\subsection{Comparison with an Instruction Tuned Model}
We evaluated our method against an instruction-tuned model on the MIMIC-CXR dataset, i.e., MedGemma-4B-it \cite{sellergren2025medgemma}, to demonstrate the superiority of our approach over instruction-tuned baselines.
MedGemma-4B was fine-tuned on the MIMIC-CXR dataset and showed strong performance in report generation and classification of five diseases: atelectasis, cardiomegaly, consolidation, edema, and pleural effusion.
Also, we compared our method with LLaMA-3.2-11B-Vision-Instruct \cite{dubey2024llama} as a baseline without fine-tuning.
Since reports in the MIMIC-CXR dataset consist of findings and impression sections, we perform response shuffling between these sections without using a separator.

\begin{table}[t]
\centering
\caption{VQA accuracy on the MIMIC-CXR dataset. We compared our method with MedGemma-4B \cite{sellergren2025medgemma}, an instruction-tuned model trained on the MIMIC-CXR dataset, and LLaMA-3.2-11B-Vision \cite{dubey2024llama}.}
\label{tab_mimic}
\begin{tabular}{l|l}
\hline
Model & Accuracy \\
\hline
LLaMA-3.2-11B-Vision (w/o FT) & 45.4 \\
MedGemma-4B & 40.0 \\
InstFree & 55.4 \SD{0.38} \\
InstFree \w RS & \textbf{55.6} \SD{0.19} \\
\hline
\end{tabular}
\end{table}

Table \ref{tab_mimic} shows multiple-choice VQA accuracy on the MIMIC-CXR dataset (i.e., Medical-CXR-VQA \cite{hu2024interpretable}), and Fig. \ref{fig_radar}(c) presents a radar chart of the accuracy for each question type.
InstFree and InstFree \w RS achieve higher accuracy than MedGemma-4B across all question types, suggesting that our approach provides a practical alternative to instruction tuning for developing an LVLM tailored to the target dataset.
MedGemma-4B achieves relatively high accuracy on presence-type questions, while exhibiting lower accuracy on other question types.
We believe that the low accuracy across most MIMIC-CXR question types is due to limited exposure to diverse instructions during MedGemma-4B’s training.
By contrast, presence-type questions beginning with ``Is there'' align closely with MedGemma-4B’s chest X-ray classification prompts and consequently yield higher accuracy.
Moreover, the low accuracy on abnormality-type questions likely stems from the much larger pool of 281 answer candidates in MIMIC-CXR, compared with the five diseases considered in the MedGemma-4B evaluation.
These observations suggest that discrepancies between training and evaluation instructions can significantly impact model performance, highlighting the challenges and limitations of constructing an instruction dataset that covers the full range of medical problems.

\noindent\textbf{Discussion.} Medical instruction tuning typically relies on labor-intensive instruction data \cite{fleming2024medalign} (e.g., even a relatively small dataset with 3,515 question-answer pairs for 315 images required approximately 60 hours of specialist contributions from 15 clinical trainees, in addition to expert validation \cite{lau2018dataset}).
In contrast, our approach outperforms MedGemma-4B using only existing image-description pairs, without any cost for constructing instruction datasets.
Furthermore, whereas MedGemma is developed through large-scale multi-stage training (vision encoder adaptation, multimodal pretraining, and post-training via distillation and reinforcement learning) \cite{sellergren2025medgemma}, our approach requires a total training time of only about 4, 5, and 5 hours on a single GPU for WBCAtt, SKINCON, and CBIS, respectively, and about 129 hours for the larger MIMIC-CXR, including the warm-up stage.
In addition, the proxy instruction adds only eight learnable vectors, which are negligible compared with the 11B parameters of the LVLM.
Since it is discarded at inference, the deployed model retains the same architecture and inference cost as the original LVLM.

\noindent\textbf{Response shuffling on free-text reports.} Unlike on SKINCON, WBCAtt, and CBIS, whose descriptions consist of clearly separable attributes, response shuffling yields only a marginal gain on MIMIC-CXR.
This is because the impression section often duplicates or summarizes the findings section, which naturally limits the additional gain from shuffling.
Therefore, fully exploiting response shuffling on real-world medical texts may require designing clinical attributes and extracting them from the texts, which we leave for future work.

\subsection{Ablation Studies}

\begin{table}[t]
\centering
\caption{Ablation study of the proxy instruction on the SKINCON, WBCAtt, and CBIS datasets. A dash (--) denotes fine-tuning with a text instruction instead of a proxy instruction, which is identical to LLaMA-3.2-11B-Vision (FT). Fix and Update denote InstFree variants fine-tuned with fixed and continuously updating proxy instructions, respectively. Gradual denotes our method, which fine-tunes the model with the gradual proxy update.}
\label{tab_abl}
\begin{tabular}{l|lll|l}
\hline
Proxy & SKINCON & WBCAtt & CBIS & Avg. \\
\hline
-- & 59.4 \SD{3.85} & 61.7 \SD{3.72} & 63.9 \SD{0.3} & 61.6 \\
Fix & 63.9 \SD{1.52} & 65.1 \SD{0.97} & 61.7 \SD{1.49} & 63.5 \\
Update & 66.4 \SD{2.55} & 63.6 \SD{0.83} & 63.7 \SD{0.49} & 64.5 \\
Gradual & \textbf{69.8} \SD{0.75} & \textbf{68.4} \SD{0.61} & \textbf{65.0} \SD{0.37} & \textbf{67.7} \\
\hline
\end{tabular}
\end{table}

\subsubsection{Comparison with Instruction-Free Tuning Variants}
We compared our method with two instruction-free tuning variants, i.e., Fix and Update, to demonstrate the effectiveness of the proposed gradual proxy update.
To isolate the effect of the proxy instruction, we excluded response shuffling from all models in this comparison.
Fix, analogous to ProMoT \cite{wang2024two}, used a fixed proxy instruction derived from a prior warm-up stage and updated only the vision encoder during fine-tuning, whereas Update referred to fine-tuning in which both the proxy instruction and the vision encoder were updated via gradient descent.

Table \ref{tab_abl} shows the multiple-choice VQA accuracy of instruction-free tuning variants.
All models fine-tuned with a proxy instruction achieve higher average accuracy than the model fine-tuned with a text instruction (--), demonstrating the effectiveness of the proxy instruction.
Among the variants, accuracy differs between the Fix and Update, as Fix achieves higher performance on WBCAtt, while Update shows higher accuracy on SKINCON and CBIS.
The superior accuracy of WBCAtt with Fix might be due to the vision encoder having fewer changes during fine-tuning, probably because the cells are always centered, which leads to more consistent image features.
Notably, employing the gradual proxy update outperforms Fix by more than 4\%, confirming the benefits of capturing overall fine-tuning trends while prioritizing updates to the vision encoder parameters.

\subsubsection{Ablation on Response Shuffling}
We conducted ablation studies on response shuffling using the SKINCON, WBCAtt, and CBIS datasets.
First, to isolate the contribution of response shuffling, we compared InstFree and FT (i.e., LLaMA-3.2-11B-Vision fine-tuned with a text instruction) with and without response shuffling.
Second, we compared InstFree \w Balance (balanced sampling) to investigate whether the issue originates from the model overfitting to previous word correlations or recurring response patterns.
We calculate the frequency of each word and assign higher selection probabilities to samples that contain words with lower overall frequencies.
Third, we compared InstFree \w RS (``~ ~'') to demonstrate that accuracy degrades when response shuffling uses incorrect separators such as `` '' (instead of the ``,'' separator).
Lastly, we compared FT \w Random (fine-tuning LLaMA-3.2-11B-Vision with randomized text instructions) to demonstrate that diversifying instructions does not lead to better performance (as an extension of response shuffling).
To do this, we randomly select an instruction from a set of 58 predefined text instructions, such as \emph{``Please report this medical scan.''}. 

\begin{table}[t]
\centering
\caption{Ablation study of response shuffling on the SKINCON, WBCAtt, and CBIS datasets. A dash (--) denotes the model without any strategy. RS denotes response shuffling, and RS (``~ ~'') denotes response shuffling with an incorrect separator. Balance denotes balanced sampling, and Random denotes fine-tuning with randomized text instructions.}
\label{tab_rs}
\begin{tabular}{l|l|lll|l}
\hline
Method & Strategy & SKINCON & WBCAtt & CBIS & Avg. \\
\hline
\multirow{4}{*}{InstFree} & -- & 69.8 & \textbf{68.4} & 65.0 & 67.7 \\
 & Balance & 65.4 & 63.4 & 60.8 & 63.2 \\
 & RS (``~ ~'') & 67.9 & 57.9 & 67.2 & 64.3 \\
 & RS & \textbf{75.2} & 65.2 & \textbf{68.3} & \textbf{69.5} \\
\hline
\multirow{3}{*}{FT} & -- & 59.4 & 61.7 & 63.9 & 61.6 \\
 & Random & 3.1 & 27.5 & 27.9 & 19.5 \\
 & RS & 74.1 & 62.0 & 67.3 & 67.8 \\
\hline
\end{tabular}
\end{table}

Table \ref{tab_rs} shows the accuracy of the ablation study on response shuffling across the SKINCON, WBCAtt, and CBIS datasets.
Applying response shuffling improves the average accuracy of both InstFree and FT, confirming its effectiveness under both proxy and text instructions.
Although InstFree \w RS shows a decrease in accuracy on WBCAtt compared to InstFree, the advantages of our response shuffling are demonstrated in attribute-wise accuracy.
Fig. \ref{fig_radar}(a) and Fig. \ref{fig_radar}(b) show radar charts for each attribute on the WBCAtt and CBIS datasets, respectively.
While leveraging inter-attribute correlations can improve overall accuracy, over-reliance on them may compromise the reliability of certain attribute predictions.
In particular, response shuffling improves the accuracy on the label attribute of WBCAtt, indicating that it reduces over-reliance on previous words.
Similarly, on the CBIS dataset, response shuffling improves the accuracy on most attributes, suggesting that it enhances the reliability of attribute predictions.
Overall, even with response shuffling, FT remains less accurate than InstFree \w RS across all datasets, highlighting the benefits of the proxy instruction with gradual proxy update.

InstFree \w Balance shows lower accuracy than InstFree without balanced sampling or response shuffling.
These results indicate that oversampling of specific samples leads to overfitting and results in lower accuracy, suggesting that reducing over-reliance on previous words is more important.
InstFree \w RS (``~ ~'') shows lower accuracy than the model that uses a comma ``,'' as the separator.
Although standard sentence segmentation libraries are generally sufficient, domain-specific datasets often contain text with poorly structured sentences; therefore, carefully designing strategies to separate elements is necessary for effective response shuffling.
Meanwhile, FT \w Random exhibits significantly lower accuracy.
This suggests that the fine-tuned model tends to ignore the given instructions, consistently generating responses that follow the format of our medical reports (i.e., \emph{image-dependence}), such as \emph{``plaque, scale...''}
This result implies that diversifying instructions during fine-tuning degrades the LVLM’s instruction-following capability, thereby leading the model to generate responses primarily based on the visual input.

\subsubsection{Discussion of Misalignment}
As shown in Table \ref{tab_rs} (FT \w Random), fine-tuning the model to generate consistent responses across a broad range of instructions leads to a significant degradation of its pre-trained instruction-following capability.
This degradation appears to stem from a misalignment between the fine-tuning dataset and the pre-training data, as the pre-training paired outputs vary in format depending on the instruction.
Similarly, both the drift from the fixed proxy instruction during fine-tuning (see Fix in Table \ref{tab_abl}) and the misalignment between a naively paired single instruction and the pre-training dataset (see FT in Table \ref{tab_main}) can be interpreted as factors that adversely affect instruction-following capability.
Meanwhile, continuously minimizing misalignment during fine-tuning may appear ideal; however, excessive reliance on instructions (which will be discarded) through direct gradient descent can ultimately lead to a degradation in VQA performance (see Update in Table \ref{tab_abl}).
Within this context of misalignment, since most publicly available LVLMs do not release their datasets, manually replicating such datasets to reduce misalignment is impractical.
Therefore, the proxy instruction offers a practical solution for fine-tuning LVLMs without instructions.

\begin{figure}[t]
    \centering
    \begin{subfigure}{0.9\columnwidth}
        \centering
        \includegraphics[width=\linewidth]{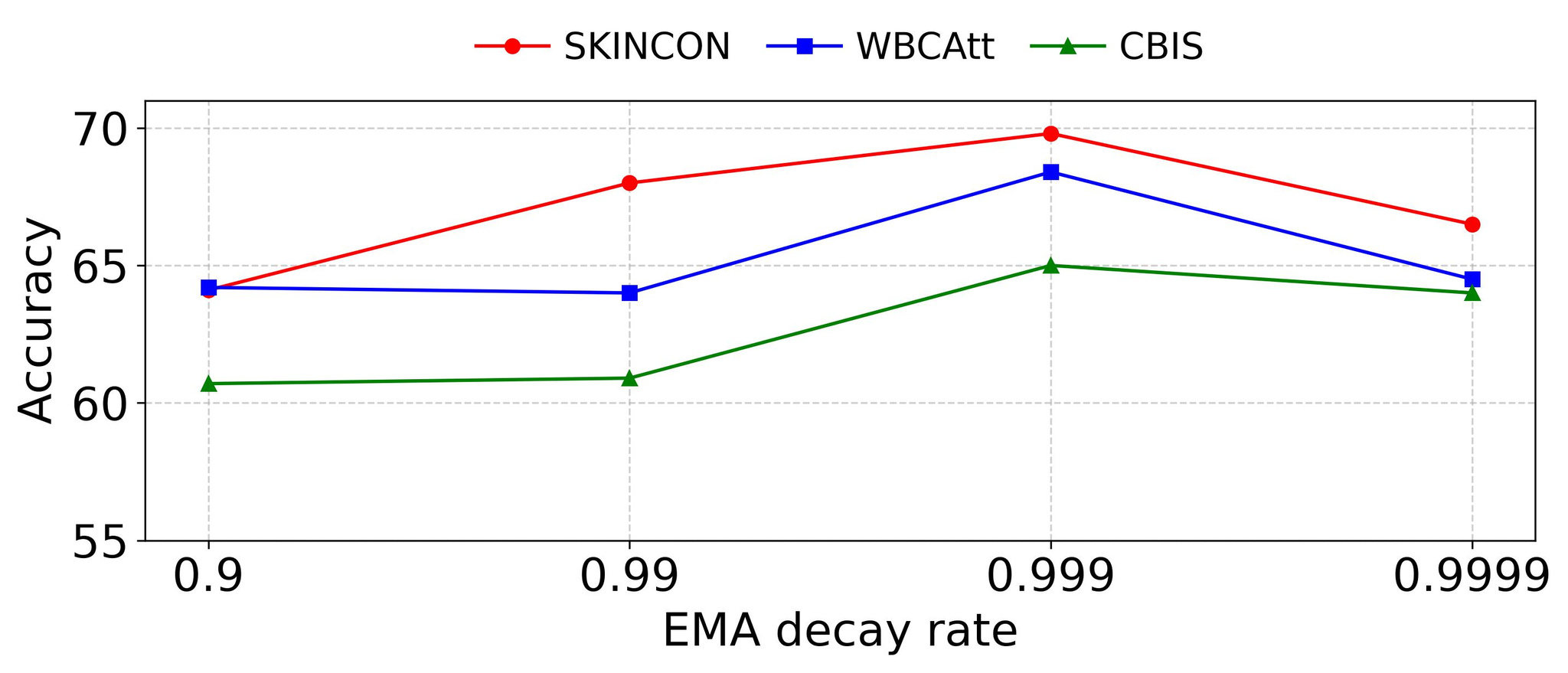}
        \caption{}
    \end{subfigure}
    \begin{subfigure}{0.9\columnwidth}
        \centering
        \includegraphics[width=\linewidth]{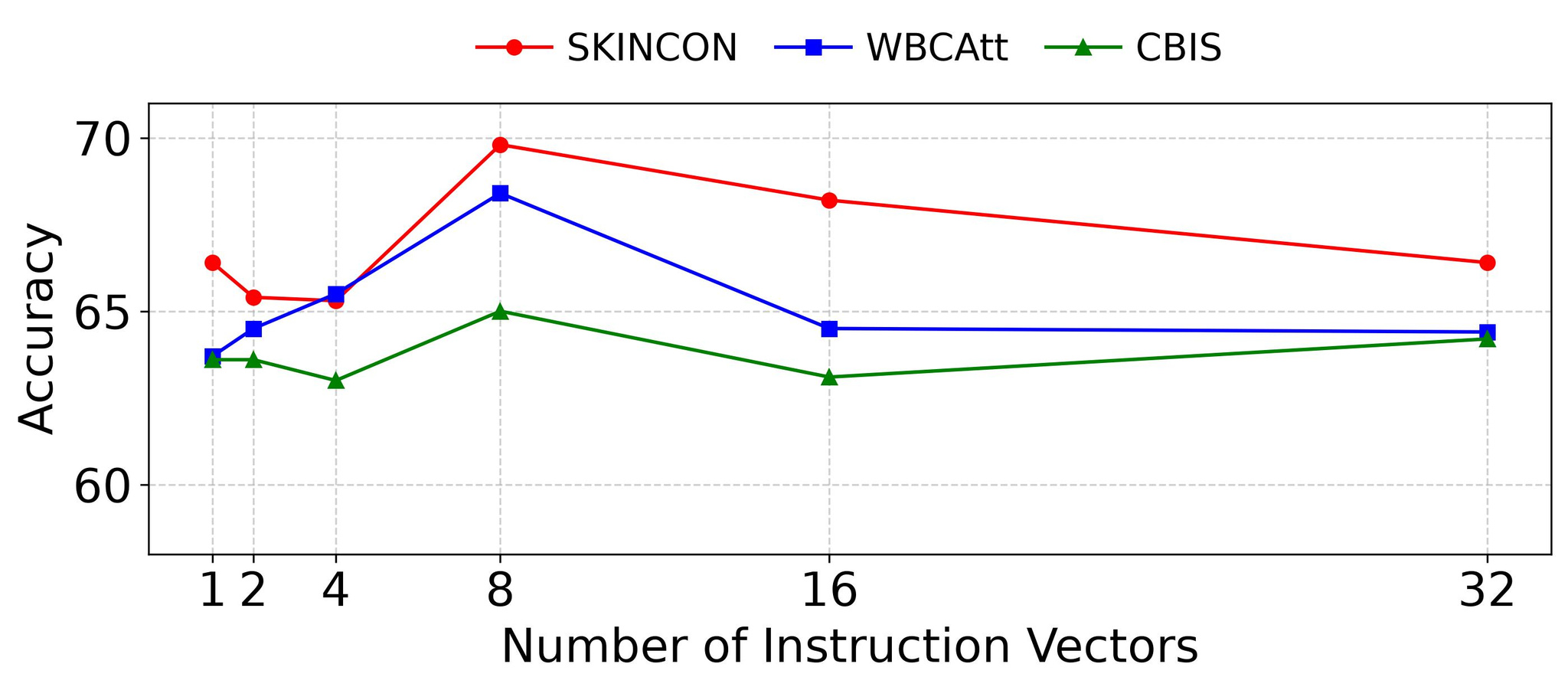}
        \caption{}
    \end{subfigure}
    \caption{Ablation study of (a) the EMA decay rate $\alpha$ and (b) the number of instruction vectors $N$ on the SKINCON, WBCAtt, and CBIS datasets.}
    \label{fig_abl}
\end{figure}

\subsubsection{Ablation on Coefficients and Instruction Scale}
We conducted ablation studies on the proxy instruction using the SKINCON, WBCAtt, and CBIS datasets by varying the EMA decay rate and the number of instruction vectors to identify optimal hyperparameters.
For the EMA decay rate $\alpha$, we experimented with values of 0.9, 0.99, 0.999, and 0.9999, while setting the number of instruction vectors $N$ to 8.
For the number of instruction vectors $N$, we experimented with values of 1, 2, 4, 8, 16, and 32, while setting the EMA decay rate $\alpha$ to 0.999.

Fig. \ref{fig_abl}(a) shows the accuracy of the ablation study on the EMA decay rate.
The value 0.999 yielded the highest accuracy, indicating that excessively fast or slow adaptation may limit performance improvements.
Fig. \ref{fig_abl}(b) shows the accuracy of the ablation study on the number of instruction vectors.
The highest accuracy was achieved with 8 instruction vectors, whereas increasing the number to 32 resulted in lower performance.
This may be related to general questions averaging about 8 tokens in length.

\begin{table}[t]
\centering
\caption{Multiple-choice VQA accuracy of Qwen2.5-VL-3B \cite{bai2025qwen2} on the SKINCON, WBCAtt, and CBIS datasets. InstFree denotes Qwen2.5-VL-3B fine-tuned using the proxy instruction with gradual proxy update.}
\label{tab_qwen}
\begin{tabular}{l|lll|l}
\hline
Method & SKINCON & WBCAtt & CBIS & Avg. \\
\hline
FT & 42.0 & 39.4 & 39.3 & 40.2 \\
InstFree & \textbf{61.3} \small($\uparrow$19.3) & \textbf{48.2} \small($\uparrow$8.8) & \textbf{49.9} \small($\uparrow$10.6) & \textbf{53.1} \\
\hline
\end{tabular}
\end{table}

\begin{figure*}[t]
    \centering
    \includegraphics[width=0.9\linewidth]{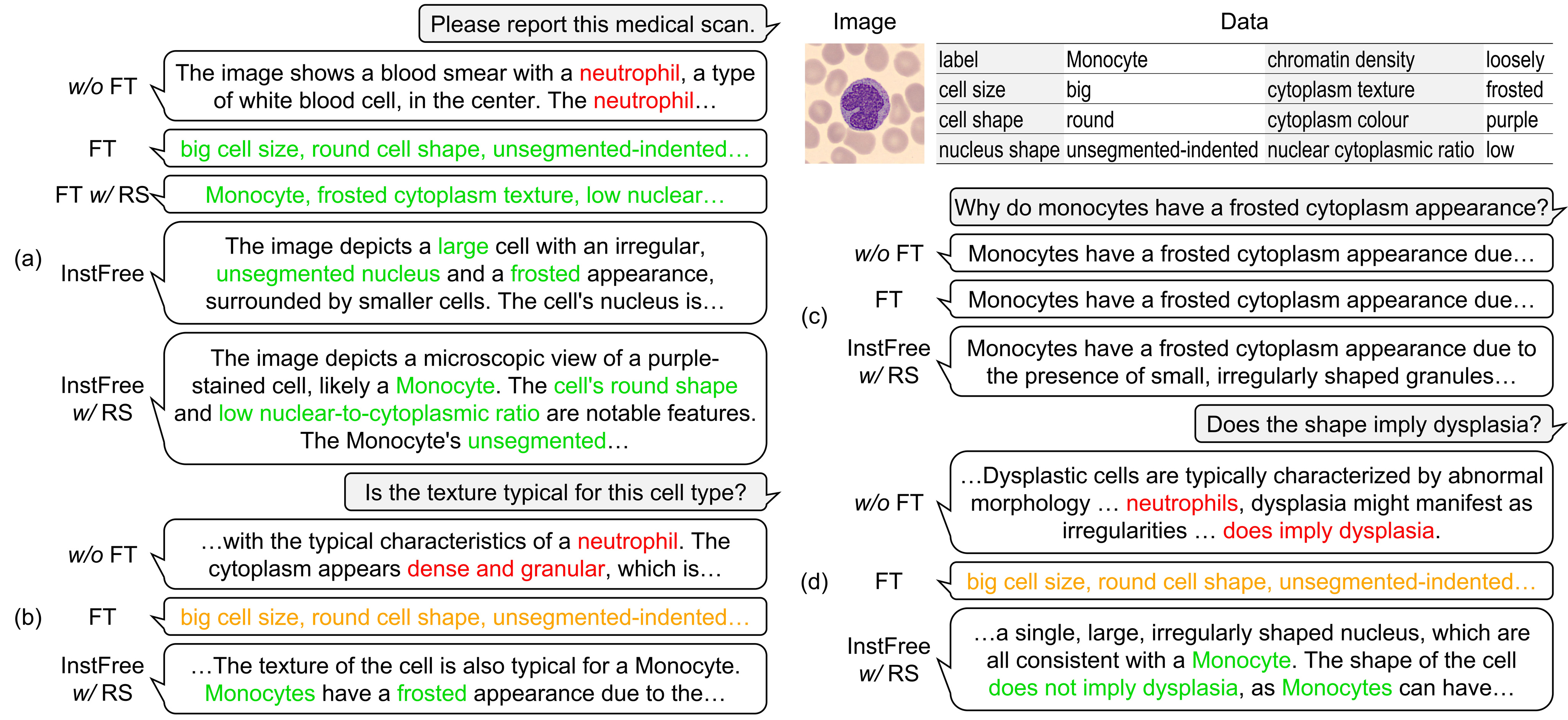}
    \caption{Qualitative VQA results on the WBCAtt dataset. Correct responses are marked in green, incorrect responses are marked in red, and responses that do not follow instructions are marked in orange.}
    \label{fig_qual}
\end{figure*}

\begin{table*}[t]
\centering
\footnotesize
\caption{Qualitative VQA results on the SKINCON, WBCAtt, and CBIS datasets for the instruction \emph{``Give a short explanation of the image.''} The first row shows each image with its paired description.}
\label{tab_qual_sup}
\renewcommand{\arraystretch}{1.3}
\begin{tabular}{@{}cm{0.24\textwidth}m{0.28\textwidth}m{0.26\textwidth}@{}}
\hline
& \multicolumn{1}{c}{\textbf{SKINCON}} & \multicolumn{1}{c}{\textbf{WBCAtt}} & \multicolumn{1}{c}{\textbf{CBIS}} \\
\hline
& \begin{minipage}[c]{2.0cm}\vspace{2pt}\includegraphics[width=\linewidth]{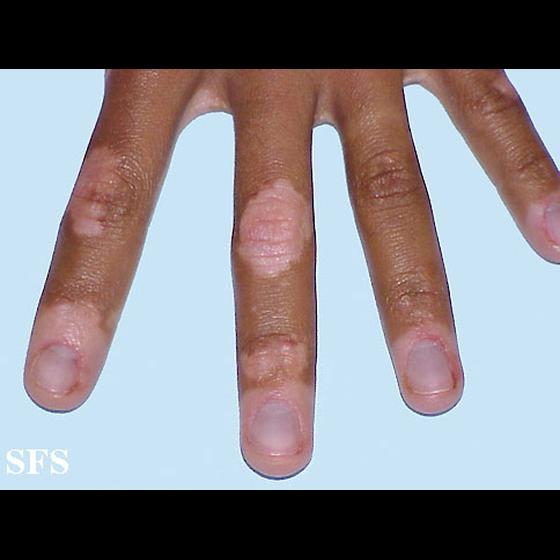}\vspace{2pt}\end{minipage}\hfill
\begin{minipage}[c]{\dimexpr0.24\textwidth-2.15cm\relax}\vspace{2pt}patch, white\newline(hypopigmentation)\vspace{2pt}\end{minipage}
& \begin{minipage}[c]{1.2cm}\vspace{2pt}\includegraphics[width=\linewidth]{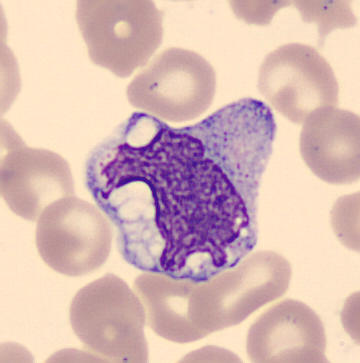}\vspace{2pt}\end{minipage}\hfill
\begin{minipage}[c]{\dimexpr0.28\textwidth-1.35cm\relax}\vspace{2pt}big cell size, irregular cell shape, unsegmented-indented nucleus shape, low nuclear cytoplasmic ratio, loosely chromatin density, cytoplasm vacuole, frosted cytoplasm texture, blue cytoplasm colour, Monocyte\vspace{2pt}\end{minipage}
& \begin{minipage}[c]{2.0cm}\vspace{2pt}\includegraphics[width=\linewidth]{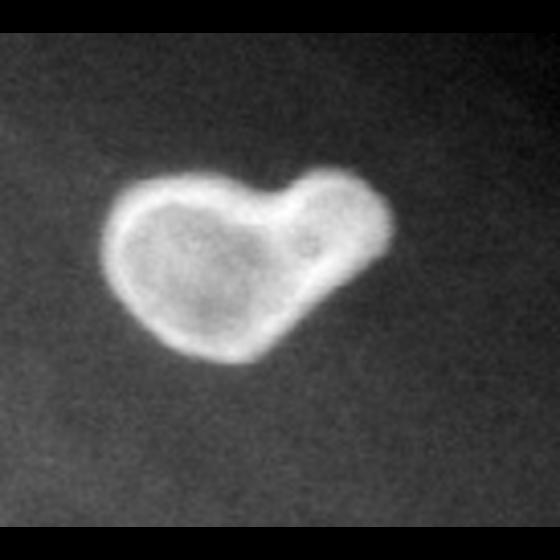}\vspace{2pt}\end{minipage}\hfill
\begin{minipage}[c]{\dimexpr0.26\textwidth-2.15cm\relax}\vspace{2pt}lucent center calc type, calcification, benign\vspace{2pt}\end{minipage} \\
\hline
\begin{tabular}{@{}c@{}}\wo FT\end{tabular}
& The image shows a hand with fingers extended, set against a light blue background. ... the skin tone is a medium brown color. ... it may be a stock photo or a medical illustration ...
& The image shows a blood smear ... including red blood cells, white blood cells, and platelets. ... White Blood Cells (WBCs): These cells are \QG{larger} and more \QG{irregularly shaped} than RBCs. ...
& The image depicts a white, irregularly shaped object against a dark background. ... Overall, the image appears to be a close-up photograph of a small, white object, possibly \QR{a piece of debris or a fragment of a larger item}. \\
\begin{tabular}{@{}c@{}}FT\end{tabular}
& \QO{patch, white(hypopigmentation)}
& \QO{big cell size, irregular cell shape, irregular nucleus shape, low nuclear cytoplasmic...}
& \QO{coarse-round and regular-lucent center calc type, calcification, benign} \\
\begin{tabular}{@{}c@{}}InstFree \\ \w RS\end{tabular}
& The image shows a hand with a \QG{white patch} on the middle finger, which is a common symptom of vitiligo.
& The image depicts a \QG{Monocyte}, a type of white blood cell, characterized by its \QG{irregular shape} and \QG{low nuclear-to-cytoplasmic ratio}. The cytoplasm is \QG{blue} and contains clear \QG{cytoplasmic vacuoles} ...
& The image depicts a \QG{calcification} of a \QG{benign} type, specifically a \QG{lucent center calcification} with a benign-appearing rim. ... Overall, the image suggests that the calcification is likely \QG{benign} in nature. \\
\hline
\end{tabular}
\end{table*}

\subsection{Generalization across LVLMs}
We further applied our method to Qwen2.5-VL-3B \cite{bai2025qwen2} to examine its generalizability to another LVLM.
Unlike LLaMA-3.2-11B-Vision, which integrates visual features through cross-attention layers, Qwen2.5-VL-3B feeds visual tokens directly into the language model. Nevertheless, as with LLaMA-3.2-11B-Vision, the proxy instruction is placed at the same position as the text instruction and updated via gradual proxy update.
We compared InstFree with FT, i.e., Qwen2.5-VL-3B fine-tuned using a text instruction.

Table \ref{tab_qwen} shows the multiple-choice VQA accuracy of Qwen2.5-VL-3B on the SKINCON, WBCAtt, and CBIS datasets.
InstFree outperforms FT on all three datasets, improving the average accuracy from 40.2\% to 53.1\%.
These results suggest that the proposed method is not limited to a specific LVLM.

\subsection{Qualitative VQA Analysis}
Fig. \ref{fig_qual} presents VQA results of LLaMA-3.2-11B-Vision on the WBCAtt dataset.
In the case of descriptive instruction (a), the model without fine-tuning (\wo FT) identifies the image as a neutrophil, whereas fine-tuned models identify it as a monocyte, demonstrating that fine-tuning allows the model to identify images more precisely.
InstFree and InstFree \w RS generate correct and natural text responses, whereas FT generates responses in the format of our medical reports.
This suggests that using text instructions similar to those employed during fine-tuning may lead to overfitting (e.g., fixed-format outputs), highlighting the advantages of our method in terms of generation quality.
Furthermore, response shuffling results in inconsistent output ordering in FT \w RS, whereas InstFree \w RS generates natural responses, demonstrating the greater robustness of our approach to output variability.

In the case of attribute-related instruction (b), InstFree \w RS accurately identifies the cell type and generates responses based on the appropriate fine-tuned attributes, whereas \wo FT misclassifies it as a neutrophil and generates responses accordingly (e.g., dense and granular; the correct is frosted).
Meanwhile, although the correct fine-tuned attributes are present in the response, FT fails to follow the instructions and generates responses that exhibit \emph{image-dependence}.
These results indicate that our method is highly sensitive to the given instructions and confirm its superior instruction-following capability.

For instruction (c), which requires medical knowledge of related attributes, all models generate natural and accurate responses, demonstrating that they retain their pre-trained knowledge even after fine-tuning.

In the case of challenging instruction (d), \wo FT identifies the cell as a neutrophil and implies dysplasia, whereas InstFree \w RS recognizes that the shape is commonly seen in monocytes and therefore does not imply dysplasia.
Meanwhile, FT generates a response that does not follow the instruction, similar to (b).
These results demonstrate that our approach enables the model to respond flexibly to diverse domain-specific instructions, without requiring expert-level human effort in dataset construction.

Table \ref{tab_qual_sup} presents additional qualitative VQA results on the SKINCON, WBCAtt, and CBIS datasets for the instruction \emph{``Give a short explanation of the image.''}
\wo FT generates generic descriptions that often lack medical accuracy (e.g., describing a calcification in a mammogram as a piece of debris), whereas FT generates fixed-format outputs regardless of the instruction (i.e., \emph{image-dependence}).
In contrast, InstFree \w RS generates natural explanations that contain the fine-tuned attributes (e.g., a white patch in SKINCON), consistent with the observations in Fig. \ref{fig_qual}.
In summary, the qualitative VQA analysis not only supports the quantitative success but also demonstrates the strengths of the proposed method beyond accuracy.

\section{Conclusion}
We introduce an instruction-free tuning framework specifically tailored for the efficient fine-tuning of medical LVLMs.
Our proposed proxy instruction preserves the instruction-following capability of the pre-trained LVLM during fine-tuning, while simultaneously promoting updates to the vision encoder.
This enables the fine-tuned model to flexibly respond to domain-specific instructions, even without explicit instructions during fine-tuning. 
Additionally, applying response shuffling during fine-tuning further mitigates the model's over-reliance on previous words, facilitating more effective fine-tuning.
Experimental results demonstrate that our method achieves superior accuracy across the SKINCON, WBCAtt, CBIS, and MIMIC-CXR datasets, significantly enhancing fine-tuning efficiency in the medical domain.
Instruction-free tuning advances medical LVLM research by enabling adaptation directly from image-description pairs, thereby bypassing a major bottleneck of instruction dataset construction.
In realistic clinical settings, this allows medical institutions to adapt LVLMs to their routinely archived data (e.g., medical images and reports), facilitating the practical and efficient use of LVLMs in healthcare.

\bibliographystyle{IEEEtran}
\bibliography{refs.bib}

\end{document}